\documentclass[letterpaper, 10 pt, conference]{ieeeconf}  

\IEEEoverridecommandlockouts                              

\usepackage{amsmath} 
\usepackage{amssymb}  
\let\labelindent\relax
\usepackage{enumitem}
\usepackage{cite}
\usepackage{booktabs}
\usepackage{multirow}
\usepackage{graphicx}
\usepackage{hyperref}
\usepackage{subcaption}
\usepackage{tabularx}
\usepackage{makecell}
\usepackage{acronym}

\acrodef{rl}[RL]{Reinforcement Learning}
\acrodef{il}[IL]{Imitation Learning}
\acrodef{dagger}[DAgger]{Dataset Aggregation}
\acrodef{pomdp}[POMDP]{Partially Observable Markov Decision Process}
\acrodef{dwa}[DWA]{Dynamic Window Approach}
\acrodef{ppo}[PPO]{Proximal Policy Optimization}
\acrodef{bc}[BC]{Behavior Cloning}
\acrodef{fov}[FoV]{Field of View}
\acrodef{amp}[AMP]{Adversarial Motion Priors}
\acrodef{cnn}[CNN]{Convolutional Neural Network}
\acrodef{mlp}[MLP]{Multi-Layer Perceptron}
\acrodef{rnn}[RNNs]{Recurrent Neural Networks}
\acrodef{ros}[ROS]{Robot Operating System}
\acrodef{mocap}[MoCap]{motion capture}

\title{\LARGE \bf
GuideWalk: Learning Unified Autonomous Navigation and Locomotion for Humanoid Robots across Versatile Terrains
}

\author{Haoxuan Han$^{1}$, Chen Chen$^{1}$, Linao Gong$^{1}$, Xin Yang$^{1}$, Hao Hu$^{1}$, Junhong Guo$^{1,2}$, \\ Zhicheng He$^{1,2}$, Yao Su$^{2}$ and Fenghua He$^{1}$
}

\begin{document}

\maketitle
\thispagestyle{empty}
\pagestyle{empty}


\begin{abstract}

Humanoid robots have achieved strong locomotion capabilities, but reliable navigation on versatile terrains remains challenging because obstacle avoidance must be coordinated with dynamically feasible motion. In this work, we present GuideWalk, a unified end-to-end framework that integrates traversability-aware navigation guidance with terrain-adaptive locomotion teacher for humanoid navigation. Specifically, we introduce a navigation module that provides explicit velocity guidance, decoupling obstacle avoidance from terrain conditions to enable robust planning across diverse environments. We propose an efficient two-stage training strategy with composite teacher distillation, where goal-directed commands and dynamically consistent actions are aggregated and distilled into a single policy. To further improve robustness, the distilled policy is refined with reinforcement learning and an auxiliary behavior cloning objective, which promotes exploration while preserving desirable teacher behaviors. Experiments demonstrate that GuideWalk achieves terrain-adaptive, goal-directed obstacle navigation across diverse terrains. Videos are available on our project website: \href{https://guide-walk.github.io/GuideWalk}{https://guide-walk.github.io/GuideWalk}.

\end{abstract}


\section{INTRODUCTION}

Humanoid locomotion has achieved remarkable progress in balance~\cite{hub}, agility~\cite{php}, and highly dynamic motion~\cite{omnixtreme}. However, deploying humanoids in cluttered real-world environments requires more than stable locomotion. A robot must simultaneously account for terrain geometry to maintain stability and surrounding obstacles to ensure collision-free navigation. Achieving robust navigation across diverse terrains while satisfying spatial constraints remains a challenging open problem.

Recent humanoid locomotion methods have extended stable motion beyond flat terrain by incorporating onboard perception, such as depth cameras~\cite{EgocentricVisionLeggedLoco, HumanoidParkourLearning, HikingInTheWild} and LiDAR~\cite{PIM, Beamdojo, AME2}, together with terrain representations such as elevation maps or depth observations. These approaches demonstrate the feasibility of terrain-aware locomotion in challenging environments. In parallel, learning-based navigation has evolved from decoupled navigation and locomotion pipelines~\cite{traditional} toward hierarchical~\cite{ABS, hierarchical_e2e_sr, hierarchical_e2e_2023} and fully end-to-end approaches~\cite{tnavrl_e2e, gallant, focusnav}. By directly mapping onboard observations to motion, end-to-end methods enable more reactive navigation without relying on explicitly separated planning and control. However, existing end-to-end approaches often require complex network architectures~\cite{focusnav} and high-dimensional spatial representations~\cite{gallant}, imposing substantial computational and GPU memory overhead during training and deployment. Moreover, jointly learning navigation and terrain-adaptive locomotion remains training-inefficient, as sparse navigation rewards provide limited supervision for learning coordinated navigation and full-body locomotion behaviors.

To address these challenges, we propose GuideWalk, a unified end-to-end framework for terrain-aware humanoid navigation and obstacle avoidance. Rather than learning these capabilities jointly through sparse task rewards starting from scratch, we introduce a composite teacher that provides complementary navigation and locomotion supervision. The student first distills navigation and locomotion behaviors from the composite teacher via \ac{dagger}, providing an effective initialization for the second training stage. It is then finetuned with \ac{rl}, where velocity tracking rewards guided by the navigation guidance provide denser navigation supervision, while an \ac{bc} objective preserves desirable teacher behaviors. The main contributions are summarized as follows:


\begin{enumerate}[leftmargin=*]

\item We present an end-to-end navigation framework that integrates supervision from navigation guidance and a locomotion teacher, enabling a unified policy for terrain-aware navigation with obstacle avoidance.

\item We propose an efficient two-stage training strategy that first distills teacher knowledge via \ac{dagger} with critic pretraining, and then refines the student via \ac{rl} finetuning with navigation-guided rewards and an auxiliary \ac{bc} loss.

\item We validate the proposed approach through simulation and real-world experiments, demonstrating its effectiveness and robustness.

\end{enumerate}

\begin{figure*}[!htbp]
    \centering
    \includegraphics[width=1.0\textwidth]{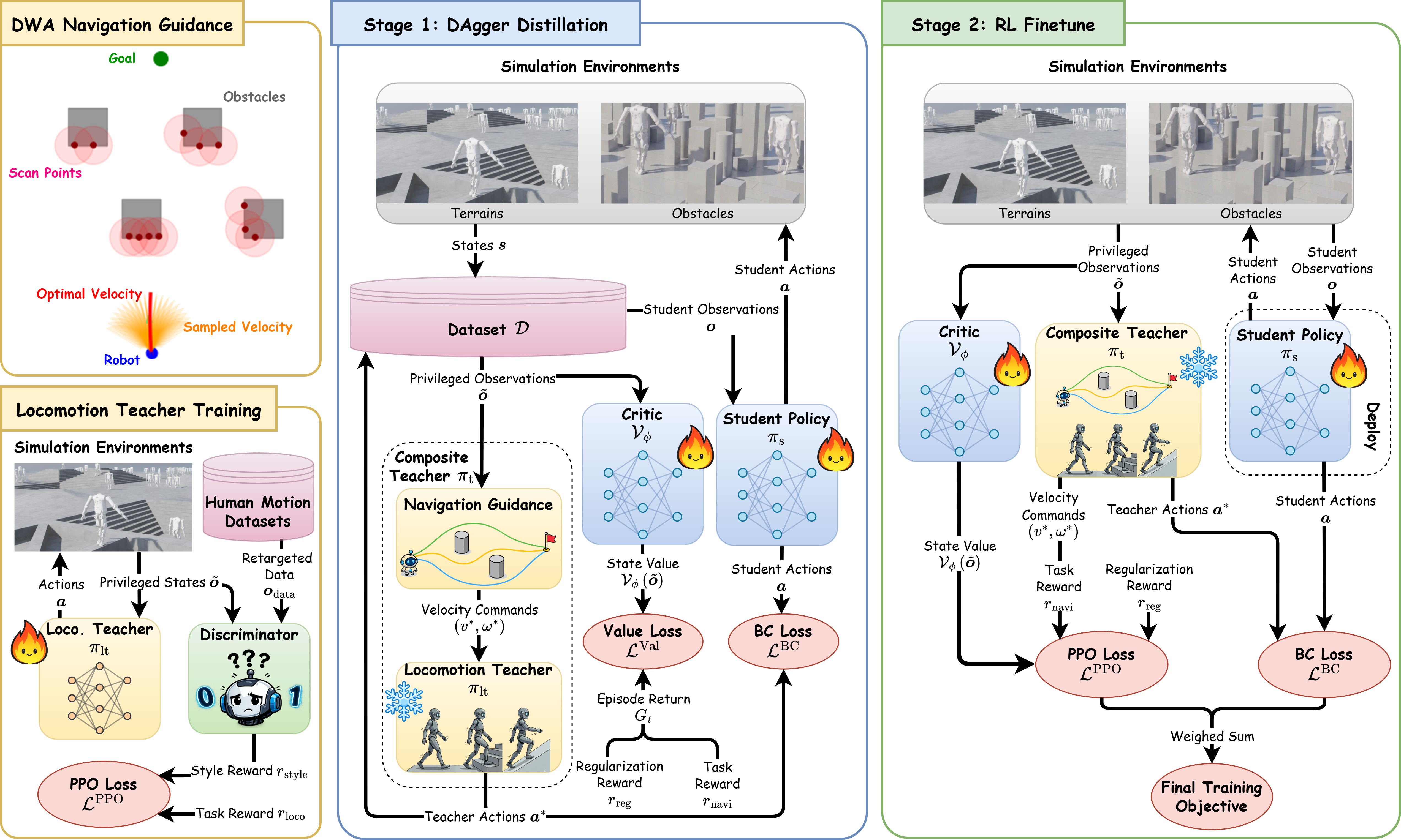}
    \caption{\textbf{Overview of GuideWalk.}
    Left: a composite teacher consisting of a DWA-based navigation module and a pre-trained locomotion policy.
    Middle: Stage 1, where the student policy and critic are trained via DAgger-based imitation from the composite teacher.
    Right: Stage 2, where reinforcement learning with an imitation objective further refines the policy.
    }
    \label{fig:overview}
\end{figure*}

\section{Methodology}

\subsection{System Overview}

The proposed framework combines guidance from a composite teacher with policy optimization via \ac{rl}, enabling the student to efficiently acquire coordinated navigation and locomotion behaviors while improving robustness and performance. As illustrated in Fig.~\ref{fig:overview}, this is achieved through a two-stage training pipeline featuring a composite teacher and a unified student policy. Specifically, the composite teacher consists of a \ac{dwa}~\cite{dwa}-based navigation guidance and a privileged locomotion teacher, jointly providing coherent supervision for high-level navigation and low-level control.

In the first stage, the student policy and critic are trained using \ac{dagger}~\cite{dagger}. Specifically, the student policy interacts with the simulation environment to collect visited states, while the composite teacher is queried to generate corresponding target actions. The student policy is optimized through supervised learning to imitate the teacher, while the critic is pretrained by regressing the returns collected from student rollouts. This stage enables the student policy to acquire both navigation and locomotion behaviors while providing a suitable value-function initialization for subsequent refinement stage.

In the second stage, the student policy is further refined using \ac{rl} based on \ac{ppo}~\cite{ppo}, with linear and angular velocity tracking rewards guided by the navigation guidance to provide dense navigation guidance. An \ac{bc} loss is incorporated alongside the \ac{rl} task objective, forming a joint optimization that improves navigaion and locomotion performance while maintaining consistency with teacher behaviors.

\subsection{Composite Teacher}

The composite teacher consists of two complementary components: a \ac{dwa}-based navigation guidance for traversability-aware velocity generation and a privileged locomotion teacher for terrain-adaptive control. The navigation guidance first generates a target velocity command $(v^*,\omega^*)$ from the target position and obstacle observations, which is then tracked by the locomotion teacher using privileged terrain and proprioceptive information. The resulting hierarchical structure provides action-level supervision that jointly captures obstacle-aware navigation and terrain-adaptive locomotion.

\subsubsection{Navigation Guidance}

The navigation guidance employs a DWA-based local planner that evaluates candidate velocity commands via trajectory rollouts and selects the optimal command according to multiple criteria. To accelerate training, it takes the robot's current pose in the global frame, the target position, and obstacle observations within the FoV as inputs, where the angular span and sensing range are designed to match the depth-camera specifications for sim-to-real fidelity. The resulting velocity command serves as the high-level locomotion target for the subsequent locomotion teacher.

To provide the navigation guidance with prior knowledge of obstacle traversability, we record the positions and geometries of obstacles during environment construction and store them in a 2D obstacle map. During training, obstacle observations are generated online by ray casting on this map, providing obstacle-only information. In contrast to the depth image, which simultaneously captures obstacles and terrain ahead of the robot, this representation distinguishes genuine obstacles from traversable terrain, such that structures such as stairs and slopes are not treated as obstacles.

To adapt the navigation speed to terrain difficulty, we use the elevation map to estimate terrain complexity and dynamically adjust the upper bound of the DWA output velocity. More challenging terrain, such as stairs, imposes a lower speed limit, whereas simpler terrain allows a higher limit.  

Although the humanoid robot possesses omnidirectional locomotion capabilities, its velocity tracking is less robust for lateral commands than for forward and yaw angular velocities. Therefore, we model the navigation guidance using a differential-drive kinematic model, restricting the output to forward linear velocity $v$ and yaw angular velocity $\omega$ to ensure reliable execution by the locomotion teacher.

For each sampled velocity pair $(v,\omega)$, future trajectories are generated via the differential-drive kinematic model, starting from the current pose and rolled out over a short horizon. The optimal velocity pair $(v^*,\omega^*)$ is chosen by maximizing the following objective:
\begin{equation}
\begin{aligned}
\arg\max_{v,\omega}\quad
&\alpha\,\mathrm{h}(v,\omega)
+\beta\,\mathrm{d}(v,\omega)
+\gamma\,\mathrm{v}(v,\omega)
+\delta\,\mathrm{c}(v,\omega) \\
\text{s.t.}\quad
&(v,\omega)\in V_r=V_s\cap V_d,
\end{aligned}
\end{equation}
where $\mathrm{h}(v,\omega)$ measures the alignment between the trajectory endpoint and the goal direction, $\mathrm{d}(v,\omega)$ measures the minimum clearance from nearby obstacles, and $\mathrm{v}(v,\omega)$ biases the selection toward higher forward speeds. $V_s$ is determined by the terrain-adaptive maximum velocity limits, while $V_d$ is determined by the prescribed acceleration limits. 

Unlike standard DWA, which treats collision avoidance as a hard constraint, we incorporate obstacle clearance as a soft constraint through $\mathrm{c}(v,\omega)$. This design avoids an empty feasible velocity set when the robot is already within an obstacle's inflation radius, allowing the guidance to select a relatively optimal velocity that moves the robot away from the obstacle and facilitates recovery.

\subsubsection{Locomotion Teacher}

The locomotion teacher maps the target velocity command and privileged terrain and proprioceptive observations to full-body joint actions. Specifically, we formulate it as a parameterized policy
\begin{equation}
\pi_{\mathrm{lt}}:
\tilde{\mathcal{O}}_{\mathrm{prop}}
\times
\tilde{\mathcal{O}}_{\mathrm{elev}}
\times
\mathcal{V}
\rightarrow
\mathcal{A},
\end{equation}
where $\tilde{\mathcal{O}}_{\mathrm{prop}}$ denotes the privileged proprioceptive observation space available during training, $\tilde{\mathcal{O}}_{\mathrm{elev}}$ denotes the clean elevation map observation space, $\mathcal{V} \subset \mathbb{R}^2$ denotes the target velocity command space, and $\mathcal{A}$ represents the full-body joint action space. Given the privileged proprioception $\tilde{\boldsymbol{o}}_{\mathrm{prop}} \in \tilde{\mathcal{O}}_{\mathrm{prop}}$, the clean elevation map $\tilde{\boldsymbol{o}}_{\mathrm{elev}} \in \tilde{\mathcal{O}}_{\mathrm{elev}}$, and the target velocity command $(v^*,\omega^*) \in \mathcal{V}$ generated by the navigation guidance, the locomotion teacher outputs the full-body joint action $\boldsymbol{a}^* \in \mathcal{A}$.

Locomotion teacher is trained via \ac{rl} with \ac{amp}~\cite{amp}, where the reference behaviors are derived from our self-collected \ac{mocap} data. This approach promotes human-like locomotion while reducing reliance on hand-crafted, terrain-specific reward shaping, which is difficult to tune consistently across diverse terrains.

Together, the navigation guidance and locomotion teacher form the composite teacher policy. Let $\mathcal{O}_{\mathrm{tar}}$ and $\tilde{\mathcal{O}}_{\mathrm{scan}}$ denote the target position space and forward 2D scan space, respectively, with $\boldsymbol{o}_{\mathrm{tar}} \in \mathcal{O}_{\mathrm{tar}}$ and $\tilde{\boldsymbol{o}}_{\mathrm{scan}} \in \tilde{\mathcal{O}}_{\mathrm{scan}}$. The navigation guidance first computes
\begin{equation}
(v^*,\omega^*)
=
\mathrm{DWA}
\left(
\boldsymbol{o}_{\mathrm{tar}},
\tilde{\boldsymbol{o}}_{\mathrm{scan}}
\right)
\in
\mathcal{V},
\end{equation}
which is then passed to the locomotion teacher together with $\tilde{\boldsymbol{o}}_{\mathrm{prop}}$ and $\tilde{\boldsymbol{o}}_{\mathrm{elev}}$. Thus, the composite teacher is written as
\begin{equation}
\pi_{\mathrm{t}}(\tilde{\boldsymbol{o}})
=
\pi_{\mathrm{lt}}
\left(
\tilde{\boldsymbol{o}}_{\mathrm{prop}},
\tilde{\boldsymbol{o}}_{\mathrm{elev}},
\mathrm{DWA}
\left(
\boldsymbol{o}_{\mathrm{tar}},
\tilde{\boldsymbol{o}}_{\mathrm{scan}}
\right)
\right),
\end{equation}
where $\tilde{\boldsymbol{o}} \in \tilde{\mathcal{O}}$ denotes the privileged observation, with
\begin{equation}
\tilde{\mathcal{O}}
=
\tilde{\mathcal{O}}_{\mathrm{prop}}
\times
\tilde{\mathcal{O}}_{\mathrm{elev}}
\times
\mathcal{O}_{\mathrm{tar}}
\times
\tilde{\mathcal{O}}_{\mathrm{scan}}.
\end{equation}
This hierarchical composition provides a teacher that jointly captures obstacle-aware navigation and terrain-adaptive locomotion for subsequent policy distillation.

\begin{table}[t]
    \centering
    \caption{Observation spaces of different modules during training and deployment.}
    \label{tab:observation_space}
    \resizebox{\columnwidth}{!}{%
    \begin{tabular}{lcccc}
        \toprule
        \textbf{Observation} 
        & \textbf{Student} 
        & \textbf{DWA} 
        & \textbf{Loco. Teacher} 
        & \textbf{Critic} \\
        \midrule
        Target position
        & \checkmark & \checkmark & & \checkmark \\
        DWA velocity command
        & & & \checkmark & \checkmark \\
        \midrule
        Noisy depth image
        & \checkmark & & &  \\
        Clean depth image
        & & & & \checkmark \\
        Forward 2D scan
        & & \checkmark & & \\
        Noisy elevation map
        & \checkmark & & & \\
        Clean elevation map
        & & & \checkmark & \checkmark \\
        \midrule
        Proprioception
        & \checkmark & & \checkmark & \checkmark \\
        Privileged proprioception
        & & & \checkmark & \checkmark \\
        \bottomrule
    \end{tabular}%
    }
\end{table}

\subsection{Student Policy}

We define the student policy as a mapping from the deployable observation space to the full-body joint action space: $\pi_\mathrm{s}: \mathcal{O}\rightarrow\mathcal{A}$, where
\begin{equation}
\mathcal{O}
=
\mathcal{O}_{\mathrm{prop}}^{5}
\times
\mathcal{O}_{\mathrm{elev}}
\times
\mathcal{O}_{\mathrm{tar}}
\times
\mathcal{O}_{\mathrm{depth}},
\end{equation}
with $\mathcal{O}_{\mathrm{depth}}$ and $\mathcal{O}_{\mathrm{elev}}$ denoting the noisy depth image and elevation map spaces, respectively, and $\mathcal{O}_{\mathrm{prop}}$ denoting the non-privileged proprioceptive observation space. Here, $\mathcal{O}_{\mathrm{prop}}^{5}$ denotes the history of five consecutive proprioceptive observations, while the depth image, elevation map, and target position are provided at the current time step. We use proprioceptive history to provide temporal cues that facilitate the distillation of locomotion-relevant latent information from the privileged teacher~\cite{history}. All observation spaces are summarized in Tab.~\ref{tab:observation_space}. All components of $\mathcal{O}$ are available during deployment.

The student policy consists of two independent \ac{cnn}s followed by a shared \ac{mlp}. The two \ac{cnn}s encode the elevation map and depth image separately, and their features are concatenated with the proprioceptive history and target position. The resulting representation is processed by the \ac{mlp} to produce the joint action.



\begin{figure}[ht]
    \centering
    \begin{minipage}[b]{0.44\columnwidth}
        \centering
        \includegraphics[width=\linewidth]{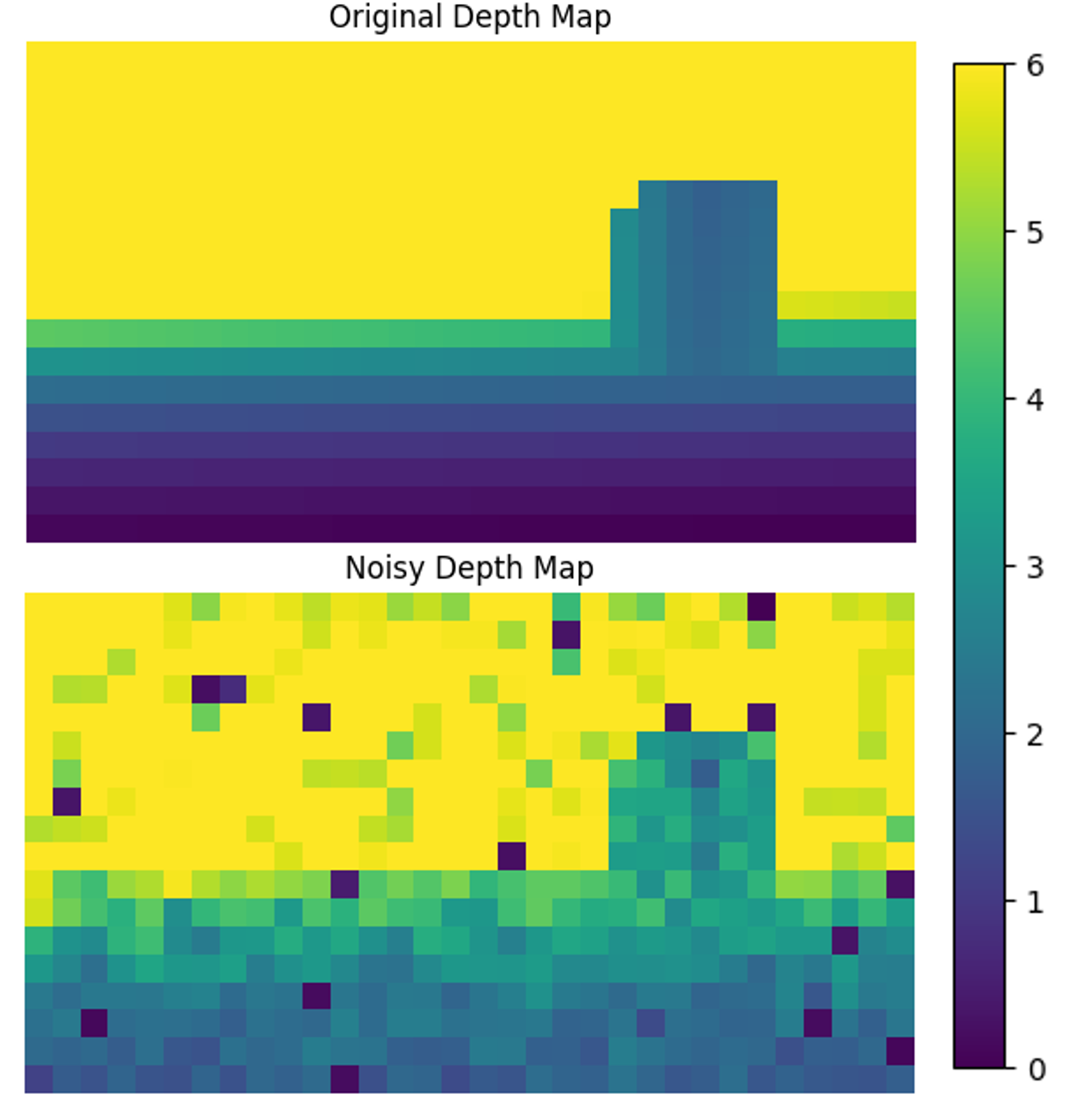}
        \captionof{figure}{Comparison of simulated depth images before and after noise injection.}
        \label{fig:noise}
    \end{minipage}
    \hfill
    \begin{minipage}[b]{0.54\columnwidth}
        \centering
        \includegraphics[width=\linewidth]{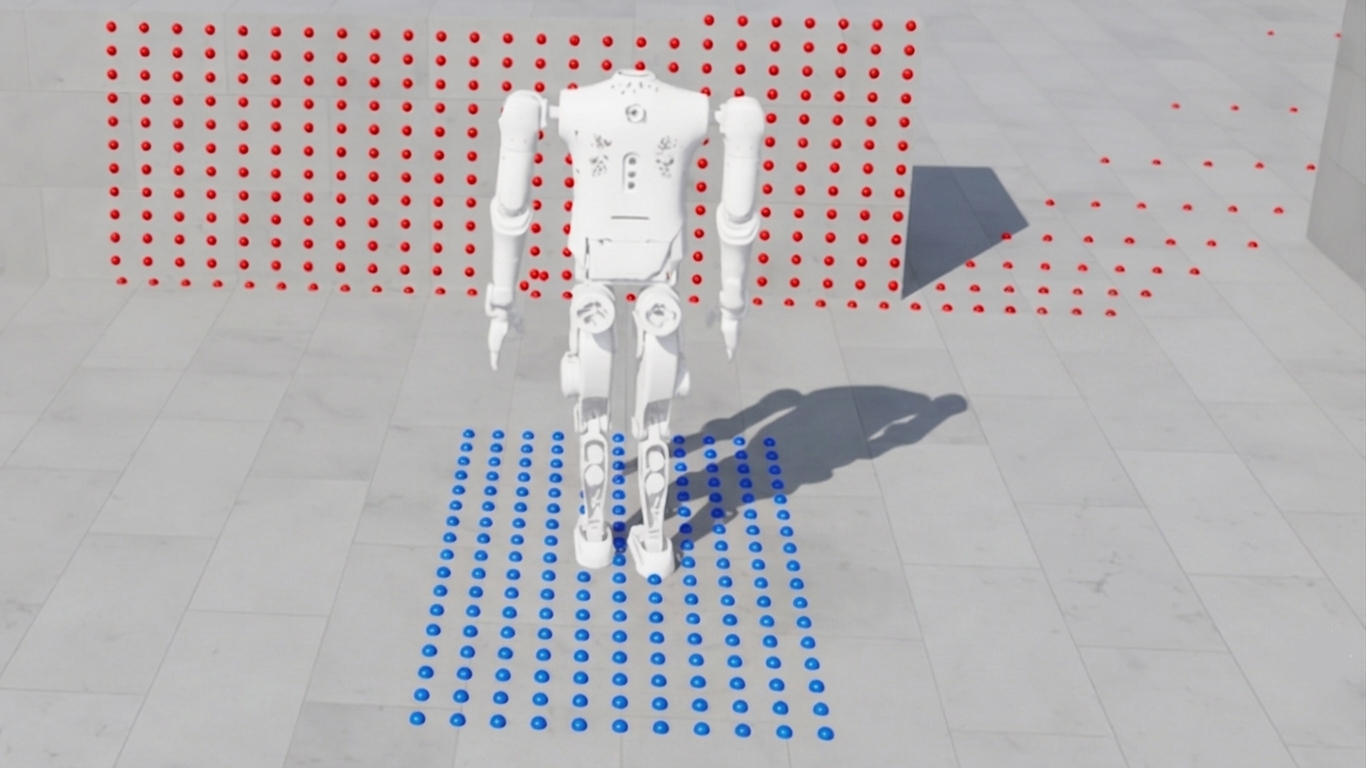}
        \captionof{figure}{Complementary perception coverage of the depth image and elevation map.}
        \label{fig:range}
    \end{minipage}
\end{figure}

To reduce the sim-to-real gap in depth perception, we augment the simulated depth images with the noise model proposed in~\cite{NowYouSeeThat}. The simulated depth images before and after noise augmentation are shown in Fig.~\ref{fig:noise}. To provide the policy with both terrain and obstacle information, the elevation map is used together with the depth image. As illustrated in Fig.~\ref{fig:range}, the elevation map primarily covers the region beneath the robot, whereas the depth image mainly covers the region directly in front of the robot, providing complementary spatial coverage for obstacle avoidance and traversing complex terrains.

\subsection{DAgger Distillation}

We employ \ac{dagger} to distill the composite teacher into the student policy. During the distillation process, the student policy $\pi_{\mathrm{s}}$ interacts with the environment and generates state trajectories according to its current policy. Let $\boldsymbol{s}_t \in \mathcal{S}$ denote the environment state at time step $t$. For the environment state $s_t$, the student receives an observation $\boldsymbol{o}_t$, while the corresponding privileged observation $\tilde{\boldsymbol{o}}_t$ is available only during training. At each visited state, the frozen composite teacher is queried using $\tilde{\boldsymbol{o}}_t$ to provide the reference action $\boldsymbol{a}_t^*=\pi_{\mathrm{t}}(\tilde{\boldsymbol{o}}_t)$. The resulting observation-action pairs ${(\boldsymbol{o}_t,\boldsymbol{a}_t^*)}$ are progressively aggregated into a growing dataset $\mathcal{D}$, which uses states visited by the student policy and reduces the distribution shift in \ac{bc}.

At iteration $i$, the student policy $\pi_{\mathrm{s},i}$ is optimized with respect to the state distribution induced by the previous policy $\pi_{\mathrm{s},i-1}$:
\begin{equation}
\pi_{\mathrm{s},i}
=
\arg\min_{\pi_{\mathrm{s}}}
\mathbb{E}_{s \sim d_{\pi_{\mathrm{s},i-1}}}
\left[
\left\|
\pi_{\mathrm{s}}\left(\boldsymbol{o}\right)
-
\pi_{\mathrm{t}}\left(\tilde{\boldsymbol{o}}\right)
\right\|^2
\right],
\end{equation}
where $d_{\pi_{\mathrm{s},i-1}}$ denotes the state visitation distribution induced by the student policy $\pi_{\mathrm{s},i-1}$. This iterative procedure progressively aligns the student policy with the composite teacher over the states encountered by the student itself, while requiring only the deployable observation $\boldsymbol{o}_t$ as input to the student policy.


To provide accurate value estimates at the beginning of Stage 2 and improve training efficiency, the critic is pretrained using the state trajectories collected by the student policy. For each visited state $\boldsymbol{s}_t$, the corresponding return $G_t$ is computed from the rewards observed along the student rollout:
\begin{equation}
G_t =
\sum_{l=0}^{H-t-1}
\gamma^l
r(\boldsymbol{s}_{t+l}, \boldsymbol{a}_{t+l}),
\end{equation}
where $H$ denotes the episode horizon and $\gamma$ is the discount factor. The critic $\mathcal{V}_{\phi}$ is then optimized to regress the value of the current student policy:
\begin{equation}
\mathcal{L}^{\mathrm{Val}}(\phi) =
\mathbb{E}_{\boldsymbol{s}_t \sim d_{\pi_{\mathrm{s},i-1}}}
\left[
\left\|
\mathcal{V}_{\phi}(\boldsymbol{s}_t)-G_t
\right\|^2
\right].
\end{equation}
Thus, the distillation stage jointly pretrains the policy and value function, providing a policy initialization through behavioral cloning and a value-function initialization consistent with the student-induced state distribution. The resulting actor and critic parameters are subsequently used to initialize the refinement stage.

\begin{table}[t]
    \centering
    \caption{Reward Terms}
    \label{tab:reward_terms}
    \begin{tabular}{l l}
        \toprule
        \textbf{Term} & \textbf{Function} \\
        \midrule
        DWA lin. vel. tracking 
        & $\exp(\|v_{xy}-v_{xy}^{\text{DWA}}\|^2)$ \\
        DWA ang. vel. tracking 
        & $\exp(\|\omega_{z}-\omega_{z}^{\text{DWA}}\|^2)$ \\
        
        \midrule
        Undesired termination 
        & $\mathbf{1}_{\text{reset}}$ \\
        Position tracking 
        & $\displaystyle\frac{1}{1+(d_{\text{goal}}/2)^2}\cdot\mathbf{1}_{t>T-2}$ \\
        Heading tracking 
        & $\displaystyle\frac{1}{1+(\Delta\psi/1)^2}\cdot\mathbf{1}_{t>T-2}$ \\
        Standing posture at goal 
        & $\|q-\bar{q}_{\text{stand}}\|_1
        \cdot\mathbf{1}_{t>T-1}
        \cdot\mathbf{1}_{d_{\text{goal}}<0.5}$ \\
        Correct-direction velocity
        & $\max\left(
        \text{ReLU}\bigl(
        \frac{v_x}{V_{\text{max}}}\cdot\mathbf{1}_{\text{dir}}
        \bigr),
        \mathbf{1}_{d_{\text{goal}}<0.5}
        \right)$ \\
        
        \midrule
        Regularization reward terms & - \\
        
        \bottomrule
    \end{tabular}
\end{table}

\subsection{RL Finetune}

To further refine the student policy $\pi_{\mathrm{s}}$ while preserving the desirable behaviors learned from the composite teacher, the second stage jointly optimizes reinforcement learning and behavioral cloning objectives. The overall objective is defined as
\begin{equation}
\mathcal{L}
=
\mathcal{L}^{\mathrm{PPO}}
+
\lambda \mathcal{L}^{\mathrm{BC}},
\end{equation}
where $\lambda$ balances task-driven policy improvement and consistency with the composite teacher.

The auxiliary \ac{bc} objective constrains the student's joint actions to remain close to those of the composite teacher, providing a behavior regularization effect. This constraint reduces the need for manually designed motion-style rewards and prevents the student from developing undesirable motions while optimizing the task reward:
\begin{equation}
\mathcal{L}^{\mathrm{BC}}(\theta)
=
\mathbb{E}_{(\boldsymbol{o},\tilde{\boldsymbol{o}})\sim\mathcal{D}}
\left[
\left\|
\pi_{\mathrm{s}}(\boldsymbol{o})
-
\pi_{\mathrm{t}}(\tilde{\boldsymbol{o}})
\right\|_2^2
\right].
\end{equation}
where $\mathcal{D}$ denotes the online trajectory buffer collected by the current student policy. 

The \ac{ppo} objective complements this behavior regularization by directly incorporating the navigation guidance into the task reward. Specifically, the linear and angular velocity tracking terms encourage the student to follow
$(v^*,\omega^*)=
\mathrm{DWA}(\boldsymbol{o}_{\mathrm{tar}},\tilde{\boldsymbol{o}}_{\mathrm{scan}})$,
providing dense navigation supervision throughout the rollout. Compared with relying solely on navigation rewards, this guidance provides more frequent learning signals and improves training efficiency. All reward terms are summarized in Table~\ref{tab:reward_terms}.

The policy parameters $\theta$ are optimized using the \ac{ppo} loss:
\begin{equation}
\mathcal{L}^{\mathrm{PPO}}(\theta)
=
\mathbb{E}
\left[
\min
\left(
r_t(\theta)A_t,
\mathrm{clip}\left(r_t(\theta),1-\epsilon,1+\epsilon\right)A_t
\right)
\right],
\end{equation}
where
$r_t(\theta)
=
\frac{
\pi_{\mathrm{s}}(\boldsymbol{a}_t\mid\boldsymbol{o}_t;\theta)
}{
\pi_{\mathrm{s,old}}(\boldsymbol{a}_t\mid\boldsymbol{o}_t)
}$
is the probability ratio and $A_t$ denotes the advantage estimate.

\begin{table}[t]
\centering
\caption{
Curriculum terrain generation parameters.
}
\label{tab:terrain_curriculum}
\small
\setlength{\tabcolsep}{4pt}
\begin{tabular}{llcc}
\toprule
Terrain type \(\tau\) & Parameter & \(p_{\tau}^{\min}\) & \(p_{\tau}^{\max}\)  \\
\midrule
\multirow{2}{*}{Obstacles} 
& Minimum obstacle spacing (m) & 2.00 & 1.00  \\
& Number of trees (-) & 5 & 40  \\
\midrule
\multirow{1}{*}{Upslope} 
& The slope angle ranges  (\(^\circ\)) & 5.00 & 25.00  \\
\midrule
\multirow{1}{*}{Downslope} 
& The slope angle ranges  (\(^\circ\)) & 5.00 & 25.00  \\
\midrule
\multirow{2}{*}{Bridge} 
& Beam width (m) & 0.40 & 0.30  \\
& Beam height (m) & 2.00 & 5.00  \\
\midrule
\multirow{2}{*}{Upstair} 
& Width of each step (m) & 0.50 & 0.30  \\
& Height of each step (m) & 0.00 & 0.20  \\
\midrule
\multirow{2}{*}{Downstair} 
& Width of each step (m) & 0.50 & 0.30  \\
& Height of each step (m) & 0.00 & 0.20  \\
\bottomrule
\end{tabular}
\end{table}

\begin{table*}[t]
\centering
\caption{Performance comparison under different ablations and settings.}
\label{tab:ablation}
\begin{tabular}{lccc|cccccc}
\toprule
\multirow{2}{*}{Method}
& \multicolumn{3}{c|}{Obstacle Navigation}
& \multicolumn{6}{c}{Terrain Traversability} \\
\cmidrule(lr){2-4}
\cmidrule(l){5-10}
& T(s) $\downarrow$
& NSR(\%) $\uparrow$
& PO(m) $\uparrow$
& Flat & Upstair & Downstair
& Upslope & Downslope & Narrow bridge \\
\midrule
w/o Nav. guidance
& 23.14$_{\pm1.8}$ & 69.9$_{\pm2.4}$ & 0.25$_{\pm0.1}$
& \textbf{100.0}$_{\pm0.0}$ & 96.2$_{\pm0.8}$ & 96.4$_{\pm0.7}$ & 96.7$_{\pm0.6}$ & 92.4$_{\pm1.3}$ & 92.3$_{\pm1.2}$ \\

w/o Loco. teacher
& 18.07$_{\pm1.3}$ & 61.5$_{\pm1.6}$ & 0.43$_{\pm0.2}$
& \textbf{100.0}$_{\pm0.0}$ & 92.7$_{\pm0.4}$ & 90.6$_{\pm0.6}$ & 96.8$_{\pm0.2}$ & 95.3$_{\pm1.0}$ & 78.5$_{\pm0.6}$ \\

w/o Stage 1
& 16.72$_{\pm1.4}$ & 11.4$_{\pm1.2}$ & 0.21$_{\pm0.1}$
& \textbf{100.0}$_{\pm0.0}$ & 96.3$_{\pm0.8}$ & 97.9$_{\pm0.5}$ & 96.5$_{\pm0.7}$
& \textbf{97.4}$_{\pm0.6}$ & 98.3$_{\pm0.5}$ \\

w/o Stage 2
& 19.09$_{\pm1.6}$ & 18.3$_{\pm1.5}$ & 0.51$_{\pm0.2}$
& 98.1$_{\pm0.8}$ & 84.9$_{\pm1.4}$ & 89.2$_{\pm1.1}$ & 96.8$_{\pm0.7}$ & 90.6$_{\pm1.2}$ & 75.1$_{\pm1.5}$ \\

Composite Teacher
& 15.93$_{\pm1.5}$ & 73.6$_{\pm2.0}$ & 0.62$_{\pm0.2}$
& \textbf{100.0}$_{\pm0.0}$ & \textbf{98.6}$_{\pm0.8}$ & 97.6$_{\pm0.5}$ & 98.1$_{\pm0.3}$ & 96.4$_{\pm0.6}$ & 99.2$_{\pm0.3}$ \\

Deployable hierarchical
& 16.14$_{\pm1.4}$ & 60.5$_{\pm2.1}$ & 0.57$_{\pm0.2}$
& \textbf{100.0}$_{\pm0.0}$ & 89.1$_{\pm1.3}$ & 82.6$_{\pm1.6}$ & 93.4$_{\pm1.0}$ & 95.1$_{\pm1.1}$ & 86.7$_{\pm1.4}$ \\

\midrule
GuideWalk
& \textbf{15.51}$_{\pm1.2}$ & \textbf{99.0}$_{\pm0.5}$ & \textbf{0.65}$_{\pm0.1}$
& \textbf{100.0}$_{\pm0.0}$ & 96.8$_{\pm1.2}$ & \textbf{98.5}$_{\pm0.4}$
& \textbf{98.8}$_{\pm0.8}$ & 96.9$_{\pm0.9}$ & \textbf{99.8}$_{\pm0.1}$ \\
\bottomrule
\end{tabular}
\end{table*}

\section{Experiments}

\subsection{Experimental Setup}

\begin{figure}[t]
    \centering
    \includegraphics[width=1.0\columnwidth]{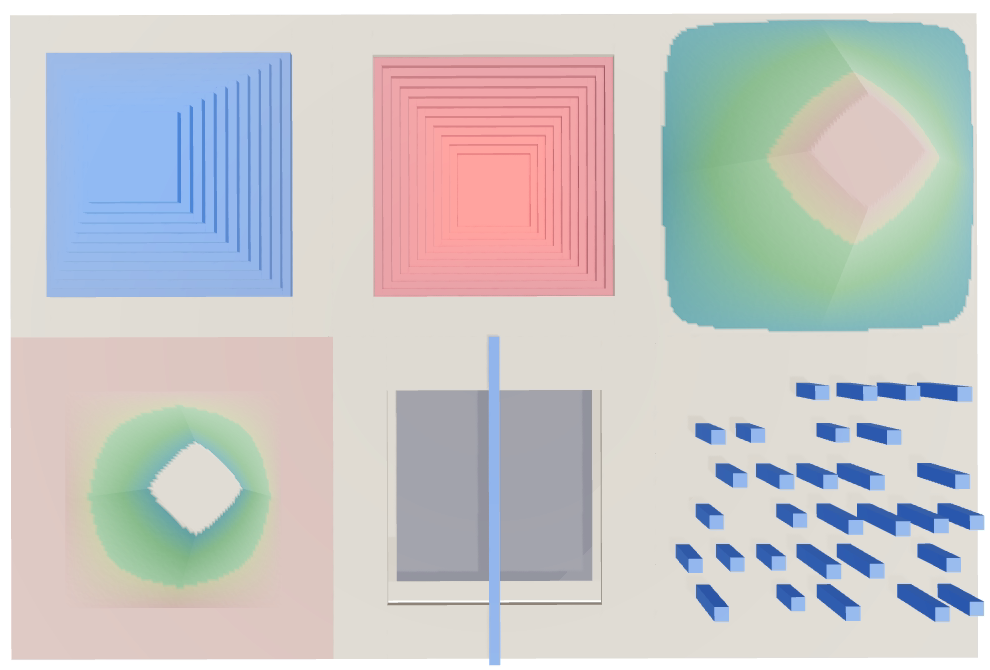}
    \caption{Basic terrain primitives and obstacle configurations used to construct training environments.}
    \label{fig:train_terrain}
\end{figure}

We train our policy in the Isaac Sim simulation platform on a single NVIDIA RTX 5080 GPU with 16 GB of memory using 2048 parallel environments. Across the two student training stages, the policy is trained for 10,000 iterations, with a total training cost of 37.7 GPU hours, including the training of the privileged locomotion teacher. The training environments, shown in Fig.~\ref{fig:train_terrain}, are constructed from basic terrain primitives, including up and down stairs, up and down slopes, and narrow bridges, while obstacle configurations can be combined with any of these terrain types, enabling terrain-aware navigation and obstacle avoidance. The curriculum terrain generation parameters are summarized in Table~\ref{tab:terrain_curriculum}. We define a difficulty scalar $s\in[0,1]$ and interpolate the terrain parameters as

$$
\mathbf{p}_{\tau}(s)
=
(1-s)\mathbf{p}_{\tau}^{\min}
+
s\mathbf{p}_{\tau}^{\max},
$$

where $\mathbf{p}_{\tau}^{\min}$ and $\mathbf{p}_{\tau}^{\max}$ denote the easiest and hardest settings, respectively. For evaluation, each terrain type is instantiated at $s=1$, corresponding to its own maximum-difficulty setting. All simulation experiments use these maximum-difficulty terrain configurations unless otherwise specified.


We conduct ablation experiments to evaluate the contribution of each component. \textit{w/o Nav. guidance} removes the navigation guidance by omitting Stage 1 and replacing the velocity command in Stage 2 with the student's body-frame linear and angular velocities from the previous timestep. \textit{w/o Loco. teacher} removes the locomotion teacher by omitting Stage 1 and the \ac{bc} loss from Stage 2, while additionally incorporating \ac{amp} to preserve the motion priors. \textit{w/o Stage 1} trains the policy with Stage 2 from scratch, whereas \textit{w/o Stage 2} retains only the distillation stage and removes the subsequent refinement. We further compare GuideWalk with the composite teacher as a reference and a deployable hierarchical baseline consisting of DWA navigation and a deployable locomotion policy obtained via \ac{amp}.

We evaluate the policy using four metrics. The \textit{navigation success rate} (NSR) measures the proportion of parallel environments in which the robot reaches within a $1\mathrm{m}$ radius of the target without collision within a given time limit. The \textit{navigation traversal time} (T) is defined as the average time required to reach the target across all successful episodes. The \textit{terrain traversal success rate} (TSR) measures the proportion of parallel environments in which the robot successfully traverses the terrain without falling within a given time limit. The \textit{Proximity to Obstacles} (PO) measures the minimum distance between the robot and surrounding obstacles along the trajectory. Specifically, NSR, T, and PO are evaluated for obstacle navigation, while TSR is used to evaluate traversal performance across different terrains.

\subsection{Main Simulation Results}

\begin{figure}[t]
    \centering
    \includegraphics[width=1.0\columnwidth]{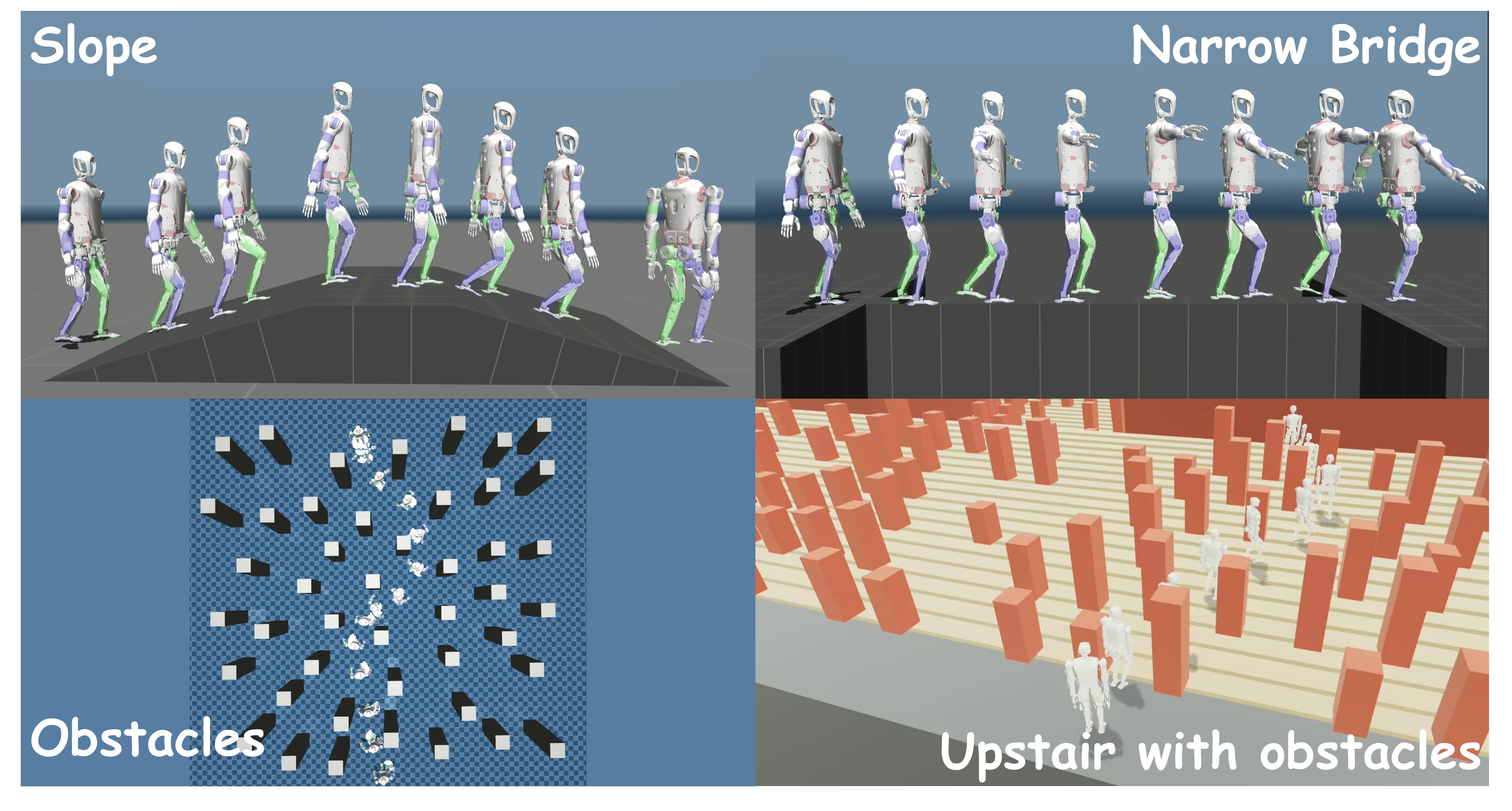}
    \caption{GuideWalk navigating diverse simulated environments.}
    \label{fig:simulation_result}
\end{figure}


As illustrated in Fig.~\ref{fig:simulation_result}, the trained policy is successfully transferred to a MuJoCo simulation environment, demonstrating its ability to perform obstacle avoidance across diverse terrains, such as navigating around obstacles on stairs.

\begin{figure}[t]
    \centering
    \begin{subfigure}[b]{0.49\columnwidth}
        \centering
        \includegraphics[width=\textwidth]{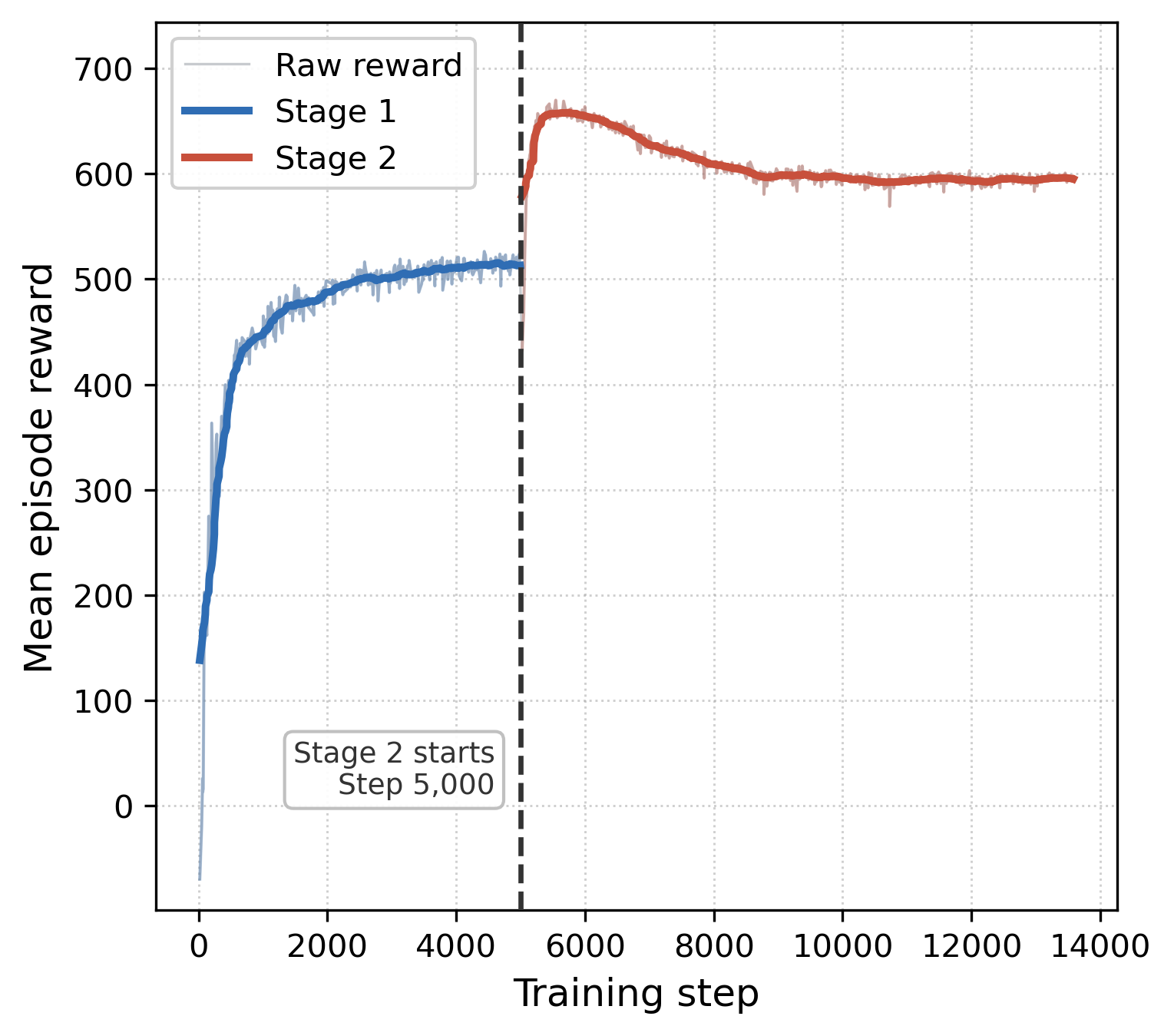}
        \caption{Mean episode reward during Stage~1 and Stage~2.}
        \label{fig:mean_reward_1}
    \end{subfigure}
    \hfill
    \begin{subfigure}[b]{0.49\columnwidth}
        \centering
        \includegraphics[width=\textwidth]{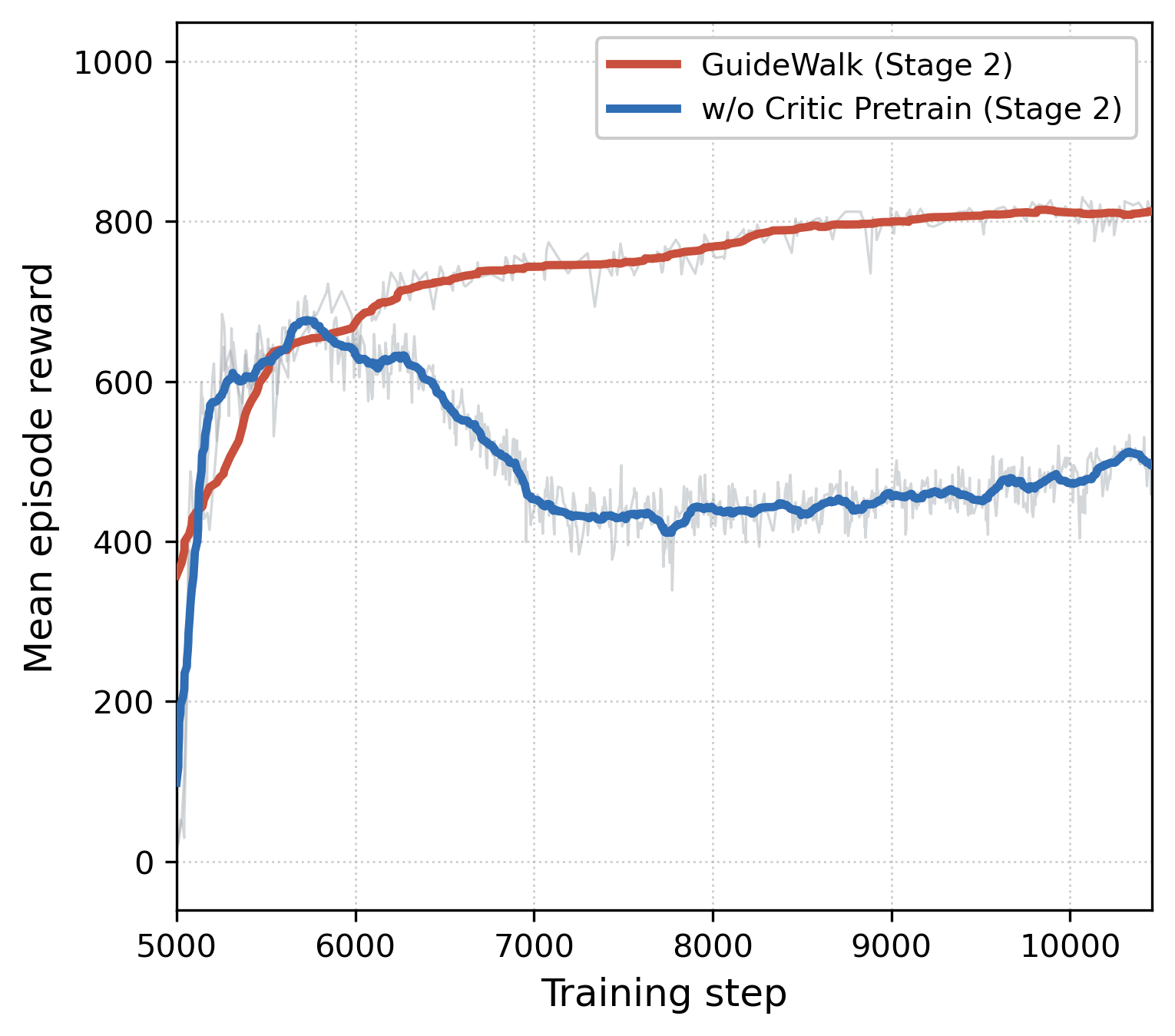}
        \caption{Stage~2 training with and without critic pretraining.}
        \label{fig:mean_reward_2}
    \end{subfigure}

    \caption{Mean episode reward curves under different settings.}
    \label{fig:mean_reward}
\end{figure}

We first compare GuideWalk with the Composite Teacher to assess whether the student preserves terrain-adaptive locomotion while improving obstacle navigation. As shown in Tab.~\ref{tab:ablation}, GuideWalk maintains comparable performance across non-flat terrains while substantially improving obstacle-navigation success, demonstrating effective transfer and further refinement of the teacher's behaviors. The training curve in Fig.~\ref{fig:mean_reward_1} shows a further increase in mean episode reward after transitioning to Stage~2.

We further compare GuideWalk with a deployable hierarchical baseline combining DWA navigation and an AMP locomotion policy. Its lower performance mainly arises from the difficulty of distinguishing traversable terrain from obstacles, which can lead to unnecessary detours or infeasible commands. In contrast, GuideWalk separates obstacle perception from terrain traversability during training, enabling unified terrain-aware obstacle avoidance with deployable observations.




\begin{figure}[t]
    \centering
    \begin{subfigure}[b]{0.3\columnwidth}
        \centering
        \includegraphics[width=\textwidth]{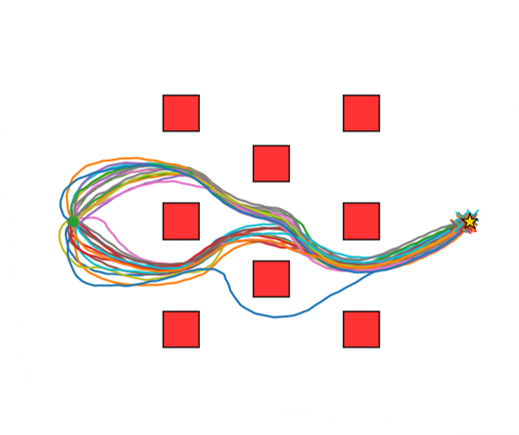}
        \caption{}
        \label{fig:traj_ours}
    \end{subfigure}
    \hfill
    \begin{subfigure}[b]{0.3\columnwidth}
        \centering
        \includegraphics[width=\textwidth]{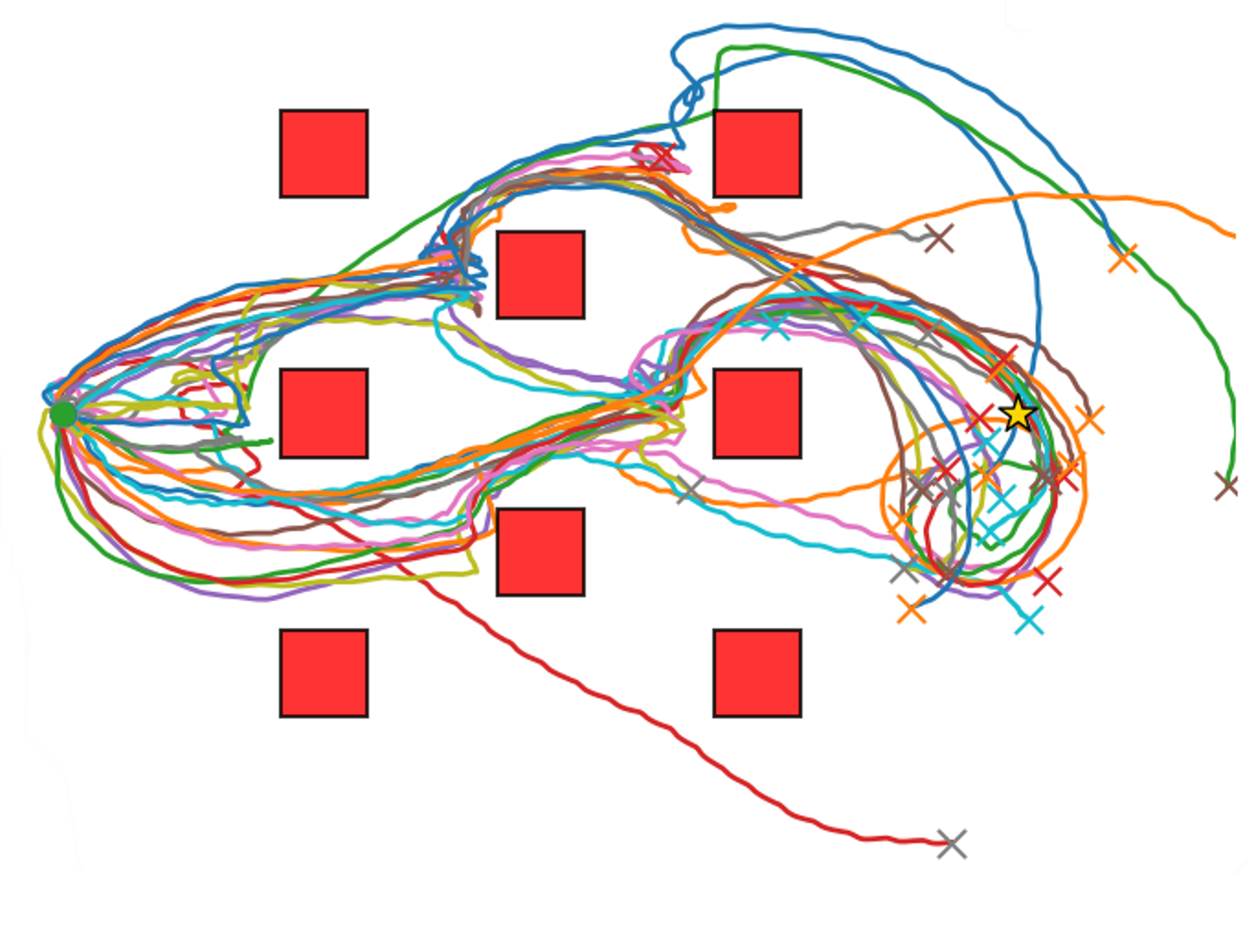}
        \caption{}
        \label{fig:traj_wo_1}
    \end{subfigure}
    \hfill
    \begin{subfigure}[b]{0.3\columnwidth}
        \centering
        \includegraphics[width=\textwidth]{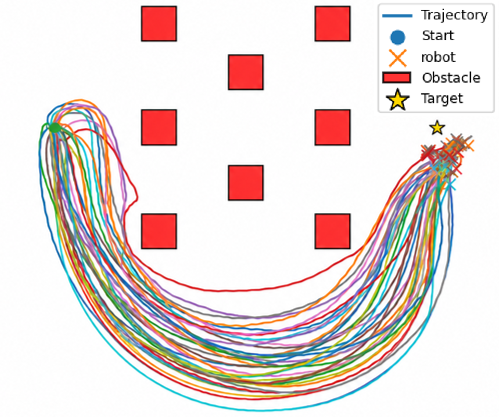}
        \caption{}
        \label{fig:traj_wo_2}
    \end{subfigure}

    \caption{Trajectory comparison under identical configurations.
    (a) GuideWalk, (b) w/o Nav. guidance, and (c) w/o Nav. guidance with increased obstacle-proximity penalty.}
    \label{fig:traj_comparison}
\end{figure}

\begin{figure}[t]
    \centering
    \begin{subfigure}[b]{0.48\columnwidth}
        \centering
        \includegraphics[width=\textwidth]{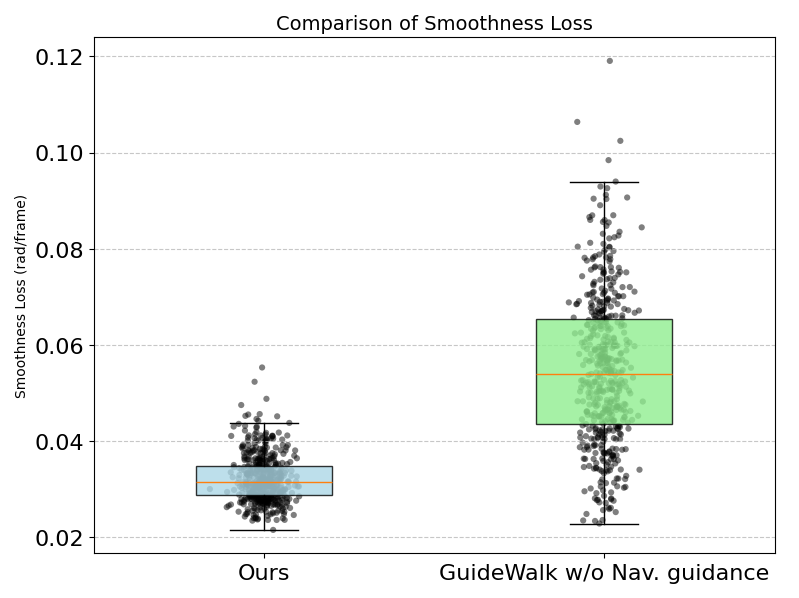}
        \caption{
        Trajectory smoothness
        }
        \label{fig:traj_smoothness}
    \end{subfigure}
    \hfill
    \begin{subfigure}[b]{0.48\columnwidth}
        \centering
        \includegraphics[width=\textwidth]{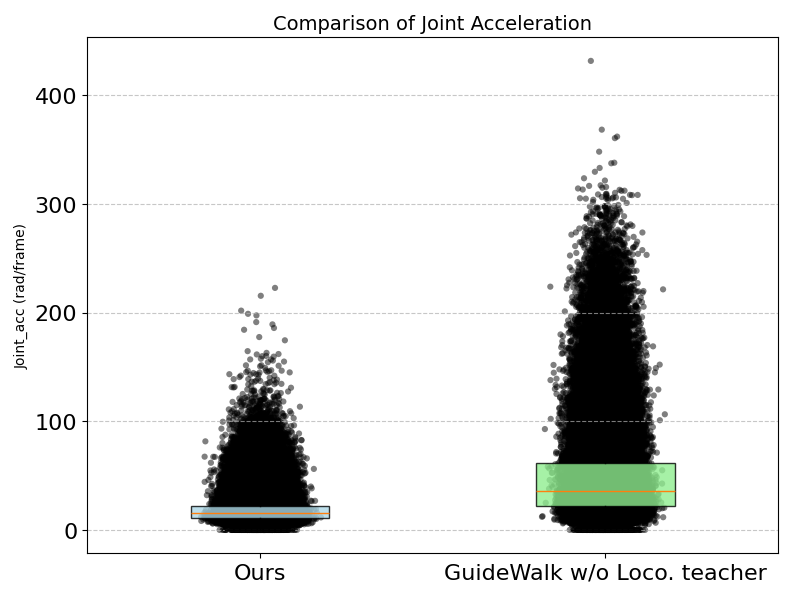}
        \caption{
        Joint acceleration
        }
        \label{fig:joint_acc}
    \end{subfigure}
    \caption{
    Motion and trajectory smoothness analysis. 
    }
    \label{fig:motion smoothness}
\end{figure}

\subsection{Ablation Studies}


\begin{figure*}[t]
    \centering
    \includegraphics[width=1.0\textwidth]{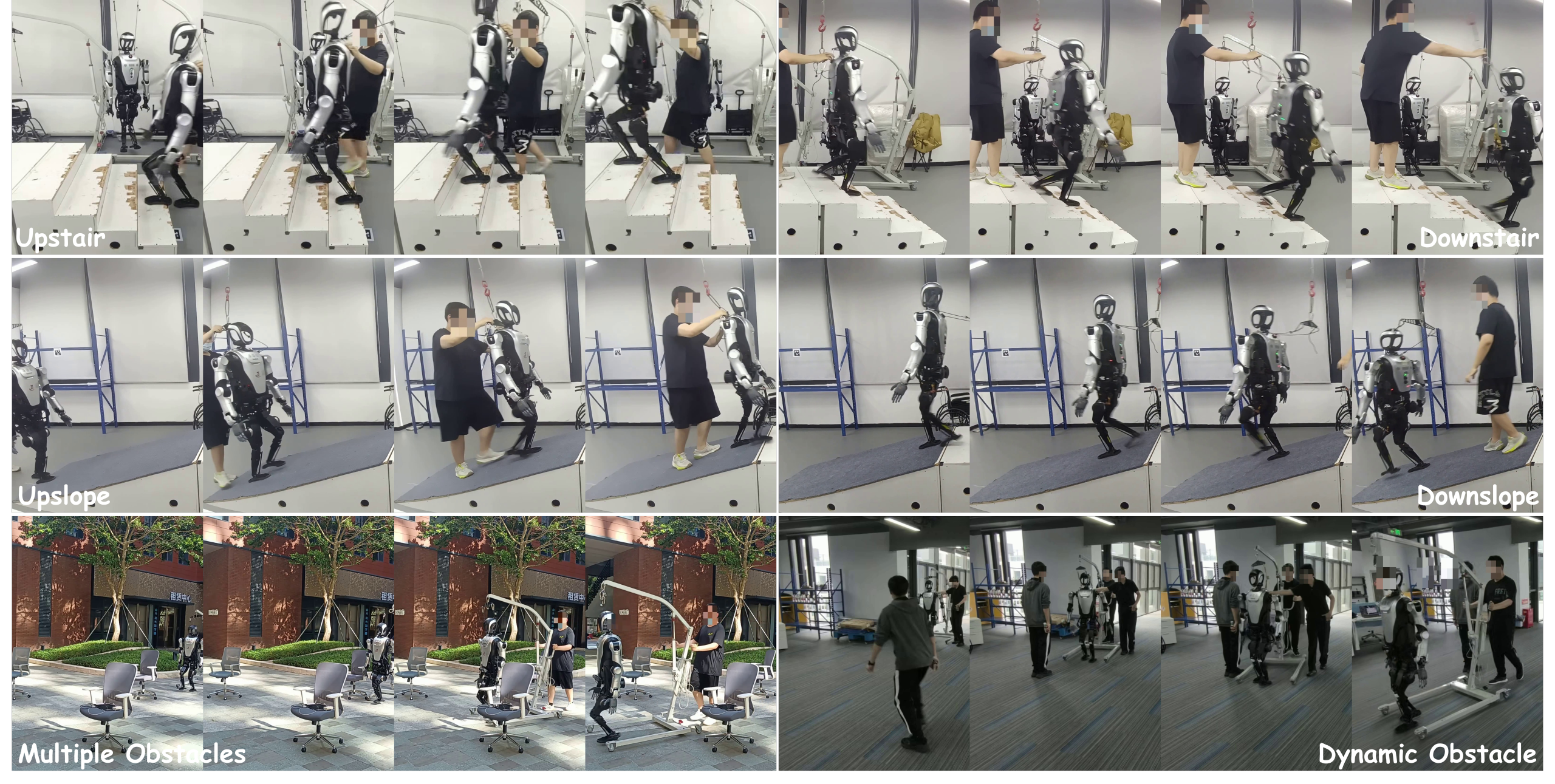}
    \caption{Real-world deployment of GuideWalk in diverse scenarios.}
    \label{fig:real}
\end{figure*}

To evaluate the role of navigation guidance, we compare GuideWalk with \textit{w/o Nav. guidance} under identical obstacle configurations. Removing the guidance causes directional oscillations and closer obstacle approaches, resulting in less smooth trajectories (Fig.~\ref{fig:traj_wo_1}, \ref{fig:traj_smoothness}). Increasing the obstacle-proximity penalty alleviates close approaches but induces unnecessary detours (Fig.~\ref{fig:traj_wo_2}), particularly with multiple obstacles. These results highlight the advantage of DWA-based guidance, which evaluates future trajectories to provide forward-looking guidance on goal alignment and obstacle clearance.

We next examine the contribution of the locomotion teacher. Removing the locomotion guidance substantially degrades terrain traversal, particularly on narrow bridges and stairs, despite retaining the AMP-based motion prior. Compared with the general motion prior provided by AMP, the locomotion teacher offers state-dependent, joint-level supervision conditioned on terrain geometry and target velocity, providing more informative guidance for terrain-adaptive whole-body control. As shown in Fig.~\ref{fig:joint_acc}, it also reduces the average joint acceleration, indicating smoother motion.

We further investigate the necessity of the two-stage training strategy. Removing Stage~1 has little effect on terrain traversal but severely degrades obstacle navigation. This indicates that obstacle navigation is highly sensitive to policy initialization and the resulting state distribution. Starting Stage~2 from an uninitialized policy leads to poor exploration of meaningful navigation states, making online teacher supervision less effective for learning long-horizon obstacle avoidance. Stage~1 via \ac{dagger} instead provides a meaningful initialization and teacher actions, enabling more effective exploration and refinement.

We further investigate the necessity of the two-stage training strategy. The complementary role of Stage~2 is evident from the \textit{w/o Stage 2} results. Without RL refinement, the distilled policy performs substantially worse in obstacle navigation, particularly in the multiple-obstacle scenario, due to the lack of task-level correction from environment feedback. Stage~2 addresses this limitation by optimizing the policy with environment feedback while preserving teacher behaviors through the auxiliary BC objective.

The auxiliary components used to stabilize Stage~2 provide complementary benefits. Without critic pretraining, the actor still inherits the navigation and locomotion behaviors from DAgger, but the randomly initialized critic produces inaccurate value estimates on the student-induced state distribution, leading to unreliable advantage estimates and unstable early PPO updates that can cause policy drift from the Stage~1 behavior. This instability is also reflected in the training curves in Fig.~\ref{fig:mean_reward_2}, where \textit{w/o Critic Pretrain} initially improves rapidly but subsequently undergoes a pronounced reward drop, whereas GuideWalk exhibits a more stable and sustained improvement throughout Stage~2. Removing the BC loss mainly affects motion quality, resulting in more irregular upper-body motions and greater deviation from the teacher. Without the joint-level imitation constraint, the policy relies more heavily on task rewards to shape its motion, making the resulting motion style more sensitive to reward design. Thus, critic pretraining stabilizes value-based policy optimization, while BC preserves desirable teacher behaviors during task-driven refinement. Together, they reduce the need for laborious reward engineering to simultaneously shape navigation performance and whole-body motion quality.


\subsection{Computational and Deployment Efficiency}

\begin{table}[t]
\centering
\caption{Comparison of computational and deployment requirements.}
\label{tab:complexity}
\footnotesize
\setlength{\tabcolsep}{3pt}
\renewcommand{\arraystretch}{1.1}
\begin{tabular}{lccc}
\toprule
Method & Network & Train GPU & Infer. Mem. \\
\midrule
Gallant~\cite{gallant}
& 3$\times$2D CNN + MLP
& 8$\times$4090 / 24 GB
& $\sim$1.3 MB \\

FocusNav~\cite{focusnav}
& \makecell{3D CNN + BEV\\+ Attn. + GRU}
& 4$\times$4090 / 24 GB
& -- \\

\textbf{GuideWalk}
& \textbf{2$\times$CNN + MLP}
& \textbf{1$\times$5080 / 16 GB}
& \textbf{$<$1 MB} \\
\bottomrule
\end{tabular}
\end{table}

To assess the efficiency of GuideWalk, we compare its computational and deployment requirements with representative humanoid navigation methods. As shown in Tab.~\ref{tab:complexity}, GuideWalk achieves its unified navigation capability with substantially lower resource requirements. This efficiency comes from both the training strategy and system design: teacher distillation provides an effective initialization before reinforcement learning, while navigation-guided velocity rewards provide dense learning signals during refinement. In addition, the simulation-friendly perception design supports efficient large-scale training, and the compact student architecture reduces computation and memory requirements for training and deployment.

\subsection{Real World Experiments}

The robot is equipped with a Livox MID360 LiDAR and an Orbbec Gemini 335Lg depth camera. On a Jetson Orin NX, we deploy FastLIO2~\cite{fast_lio} and Elevation Mapping~\cite{elevation} to provide real-time robot localization and elevation maps. The stereo RGB images are processed by Fast Foundation Stereo~\cite{ffs} to obtain depth observations at over 50~Hz, while the navigation policy is deployed on the onboard NUC for real-time inference at 50~Hz.

\begin{table}[t]
    \centering
    \caption{Real-world navigation performance.}
    \label{tab:real_performance}
    \resizebox{\columnwidth}{!}{%
    \begin{tabular}{lccccc}
    \toprule
    & Upstair & Downstair & Downslope & Upslope & Obstacles \\
    \midrule
    Success Rate
    & 7 / 10 & 10 / 10 & 10 / 10 & 10 / 10 & 10 / 10 \\
    Avg. Speed (m/s)
    & 0.40 & 0.42 & 0.48 & 0.36 & 0.56 \\
    \bottomrule
    \end{tabular}%
    }
\end{table}

We validate GuideWalk on a hardware platform, evaluating its navigation capabilities across stairs, slopes, and obstacle-rich environments, as illustrated in Fig.~\ref{fig:real}. As summarized in Tab.~\ref{tab:real_performance}, GuideWalk achieves robust real-world performance across diverse terrains, with 100\% success rates on down stairs, down slopes, up slopes and multiple-obstacle scenarios. The lower success rate on up stairs is primarily attributed to the torque limitations of the hardware platform, while the 10/10 success rate in obstacle-rich environments demonstrates reliable obstacle avoidance. The variation in average speed across terrains reflects the terrain-aware adjustment of the upper bound of the DWA velocity.

\section{CONCLUSIONS}

In this work, we presented GuideWalk, a unified end-to-end framework for autonomous navigation and locomotion of humanoid robots in diverse environments. Extensive simulation and real-world experiments demonstrate that GuideWalk effectively coordinates goal-directed navigation with stable and feasible locomotion, achieving robust performance across diverse terrains and obstacle-rich environments.

Despite these promising results, two limitations remain. First, the use of both elevation maps and depth images introduces sensing redundancy and computational overhead. Future work will explore deriving elevation maps from continuous depth observations to reduce reliance on onboard LiDAR. Second, the current policy lacks explicit spatial memory, limiting its ability to reuse information from previously observed regions. Future work will investigate incorporating spatial memory to improve navigation in more complex, partially observed environments.




\bibliographystyle{IEEEtran}
\bibliography{ref}

\end{document}